\documentclass{article}

\usepackage{amsmath}
\usepackage{graphicx}
\usepackage{natbib}
\usepackage{epstopdf}%

\usepackage{array}
\usepackage{booktabs}
\usepackage{multirow}
\usepackage{adjustbox}
\PassOptionsToPackage{hyphens}{url}
\usepackage{hyperref}
\hypersetup{
    colorlinks=true,
    urlcolor=blue
}
\usepackage{float}

\newcommand{\xml}[1]{$\texttt{<#1>}$}

\begin{document}
\label{firstpage}

\title{
    Carolina: a General Corpus of Contemporary Brazilian Portuguese
    with Provenance, Typology and Versioning Information\\{\normalsize A Preprint}}
    \author{Maria Clara Ramos Morales Crespo$^1$,\\ Maria Lina de Souza Jeannine Rocha$^1$,\\ Mariana Lourenço Sturzeneker$^1$,\\ Felipe Ribas Serras$^1$, Guilherme Lamartine de Mello$^1$,\\ Aline Silva Costa$^2$, Mayara Feliciano Palma$^1$,\\ Renata Morais Mesquita$^1$, Raquel de Paula Guets$^1$,\\ Mariana Marques da Silva$^1$, Marcelo Finger$^1$,\\ Maria Clara Paixão de Sousa$^1$,\\ Cristiane Namiuti$^3$, Vanessa Martins do Monte$^1$}

    \date{$^1$University of São Paulo, São Paulo, SP, Brazil\\
    $^2$Federal Institute of Education, Science and Technology of Bahia, Vitória da Conquista, BA, Brazil\\
    $^3$State University of Southwestern Bahia, Vitória da Conquista, BA, Brazil}


\maketitle


\section*{Abstract}
    This paper presents the first publicly available version of the Carolina Corpus and discusses its future directions. Carolina is a large open corpus of Brazilian Portuguese texts under construction using web-as-corpus methodology enhanced with provenance, typology, versioning, and text integrality. The corpus aims at being used both as a reliable source for research in Linguistics and as an important resource for Computer Science research on language models, contributing towards removing Portuguese from the set of low-resource languages. Here we present the construction of the corpus methodology, comparing it with other existing methodologies, as well as the corpus current state: Carolina's first public version has $653,322,577$ tokens, distributed over $7$ broad types. Each text is annotated with several different metadata categories in its header, which we developed using TEI annotation standards. We also present ongoing derivative works and invite NLP researchers to contribute with their own.

\section{Introduction}
\label{sec:intro}

Carolina (General Corpus of Contemporary Brazilian Portuguese with Provenance and Typology Information, ``\textit{Corpus Geral do Português Brasileiro Contemporâneo com Informações de Procedência e Tipologia}")\footnote{
  The corpus was named in honor of Carolina Michaëlis de Vasconcelos (1851-1925), a philologist who became the first woman to be a professor at a university in Portugal \citep{bassetto2015}. To know more about this choice, visit: \url{https://sites.usp.br/corpuscarolina/}.
} is a corpus of diverse typology with a robust volume of texts mainly written in Brazilian Portuguese after 1970, retrieved from the Web.

The corpus is under continuous development since September 2020 at the Digital Humanities Virtual Lab (LaViHD)\footnote{\url{https://lavihd.fflch.usp.br/}} as part of the Natural Language Processing of Portuguese Division (NLP2) of the Center for Artificial Intelligence (C4AI)\footnote{\url{https://c4ai.inova.usp.br/pt/pesquisas/\#NLP2_port}} of the University of São Paulo (USP). Currently, version 1.1 is available for download and new releases are planned for the future.\footnote{\url{https://sites.usp.br/corpuscarolina/corpus/}} With Carolina, we aim at building a robust resource with state-of-the-art features both for research in the field of Artificial Intelligence (AI) and in the field of Linguistics, focusing on the importance of provenance and a rich typology of textual information as fundamental assets in modern data availability.

The construction of Carolina is based on four concepts, which combined figure as the main difference from other available Portuguese corpora (see Section \ref{sec:related}), namely provenance, typology, versioning, and text integrality.

\textit{Textual provenance} refers to information about texts' origin and processing history \citep{werder2022provenance}. Moreover, we see provenance control as a way to intensify the curatorial aspect of building a corpus, by selecting texts with relevant content and representativeness for the purposes of the corpus and other applications, as well as ensuring that they are being distributed and used in accordance with original licenses. Easy access to textual provenance is one of the main elements for the acknowledgment of negative biases \citep{ntoutsi2020bias}. Such information allows for careful selection of texts, providing the audit trail of the data and on applications derived from it.

However, the selection of a document to a research project also involves, ideally, information about the characteristics and context of the textual piece, requiring some typological annotation about the document. This work understands \textit{typology} in a broad sense, as a crucial methodological tool for text search, selection and balancing in linguistics research. This separation has also become increasingly relevant for research in AI, which is developing models specialized in specific textual typologies, as is the case of SciBERT \citep{Beltagy2019scibert} and BioBERT \citep{lee2019biobert}, for scientific and biological texts, respectively.

These models mentioned are all pre-trained models, one of the main trends in NLP today \citep{devlin2019bert, vaswani2017transformer}. Such models require training with large volumes of complete texts. Thus, \textit{volume} and \textit{text integrality} are fundamental characteristics of a corpus that aims to allow the development of state-of-the-art algorithms for Portuguese. That would contribute to elevating that language to a state of high computational resources. With regard to linguistics research, text integrality is also an important factor, as fragmentary content can be detrimental to inter-phrase or inter-text associations in linguistics studies.

These properties require a longer process of construction, when compared to other Web corpora. Thus, \textit{versioning} becomes a practical managerial necessity for such a dataset, especially because Carolina is intended to host the textual pieces of all C4AI resources whose development is expected in the next decade, including  \textit{TaRSila} \citep{candidojr2021coraa} and  \textit{POeTiSA} \citep{pardo2021portinari}.

We present here a corpus of complete texts with annotations of provenance and typology, built according to the \textit{WaC-wiPT} methodology, that uses a strategy of versioning and continuous development to achieve, in the next years, an unprecedented volume of texts for Brazilian Portuguese corpora. The WaC-wiPT methodology was conceived based on the web-as-corpus view \citep{baroni2009wacky}, which has been dominant in recent developments in linguistic resource building \citep{fletcher2007}, with Provenance and Typology concerns, as mentioned above.

Therefore, providing an open, large and diverse corpus for Portuguese, with provenance and broad typological information, and prioritizing the use of integral or minimally modified texts, has the potential of directly impacting research both on Linguistics and Computer Science in Portuguese. This is the intended goal of this corpus, whose construction we present in this work.

\section{Related Works}
\label{sec:related}

Given the widespread availability of online content in the last decades, many researchers turned to the Web as their main source for corpus building.

The TenTen Corpus Family \citep{jakubivcek2013tenten} is available for consultation in more than 40 languages, including a corpus of 4 billion tokens for Portuguese (ptTenTen), which includes the European and Brazilian variants. The full corpus is only accessible upon payment.\footnote{\url{https://www.sketchengine.eu/}} 

OSCAR (Open Super-large Crawled Aggregated coRpus) is another multilingual corpus of considerable size, which shuffles the texts' lines to avoid copyright issues \citep[p.2]{suarez2019oscar}, with more than 18 billion words in Portuguese, without distinction of variants.

Several large corpora were built adopting the WaCky (Web-As-Corpus Kool Yinitiative) methodology, following the emergence of the first WaCky corpora: the ukWaC, deWaC, itWaC \citep{bernardini2006wacky, baroni2009wacky}, and the frWaC \citep{ferraresi2010}, which targeted English, German, Italian, and French respectively and include more than 1 billion words each. This methodology comprises four steps which are: crawling from a list of seed URLs, post-crawl cleaning, removal of duplicate content, and annotation \citep{boss2014brwac}.

One of the corpora built following this framework is the Brazilian Portuguese Web as Corpus (brWaC). Published in 2017, it contains 2.68 billion tokens that were crawled from the Web in 24 hours \citep{wagner2018brwac}. Its importance for advances in Brazilian Portuguese research in multiple areas is illustrated by its employment in NLP model training \citep{souza2020bertimbau}.

Regarding other existing Portuguese-language corpora, the virtual organization Linguateca \citep{santos2000} stands out as a center for resources focused on the computational processing of this language. Its objective is to contribute to the development of new computational and linguistic resources, facilitating the access of new researchers to existing tools. Among the available corpora at Linguateca that specifically target Brazilian Portuguese, the ones that stand out as the most significant in size are: Brazilian Corpus \citep{sardinha2010corpus}, Lácio-Web \citep{aluisio2003lacioweb}, and Corpus do Português, subcorpora NOW \citep{davies2018now} and Web/Dialects \citep{davies2016webdial}.

\section{Methodology}
\label{sec:metho}

In order to create a corpus with provenance, typology, versioning and text integrality we developed the WaC-wiPT methodology, described in detail by \cite{sturzeneker2022carolina}. In this section we will overview some of its relevant aspects, namely conception, surveys, typologies, headers and metadata. 

Concerning the methodology \textbf{conception}, we initially analyzed the pre-existing Brazilian Portuguese corpora for natural language processing, aiming at supporting the development of our methodology and considering the possibility of incorporating some of these resources into the Carolina Corpus. That enabled us to assess the benefits and drawbacks of their methodologies, given our goals. In doing so, we decided on a web-based corpus, but against the adoption of the other frameworks.

As \cite{cvrcek2020} point out when discussing the drawbacks of web-crawled corpora, the compilation of what is easily accessible on the web, although cheaper, requires from the user the effort of sorting it out, as opposed to traditional corpora which `are distinguished by careful design, determining well-motivated text categories, assigning quotas and sticking to them, even though the data might be hard to acquire' \citep[p. 2]{cvrcek2020}. For example, despite the WaCky method claiming the facilitation of an automatic balancing of content without bias, the methodology presents some drawbacks. As its own creators acknowledge, the automated methods allow for limited control over the contents that end up in the final corpus, and therefore they need to be post-hoc investigated \citep{baroni2009wacky}. 

Additionally, even the balancing of content would be limited by the nature of general web-crawling techniques, which `can only access the ``searchable" web', \footnote{As we openly distribute the contents of the Carolina Corpus online, assuring data provenance beforehand is crucial to determine the original terms of use of content. For this reason, we refrained from collecting random samples of the web also to ensure the openness of information crawled.} and most interesting content embedded in ``noisy" pages would be likely discarded by automatic filters \citep[p. 27]{cvrcek2020}.

In that sense, we aimed at building a corpus as large as other web corpora, preserving the design effort of traditional corpora. Added to the great size and curatorship of domains, this design envisions not only a greater care with copyright, but also the preservation of the features mentioned in section \ref{sec:intro}, namely conception, surveys, typologies, headers and metadata, which can often be seen in other corpora individually, but rarely in combination.

Having established the conceptualization of the methodology, the next step was to conduct \textbf{surveys}. They are in-depth research and investigation of web domains.\footnote{A non-exhaustive list of sources incorporated into Carolina is available at \url{https://sites.usp.br/corpuscarolina/repositorios/}.} The surveys started with the broad typology, divided in seven types, up to this version: judicial branch; legislative branch; datasets and other corpora; public domain works; wikis; university domains; and social media.

In addition, given that throughout the surveying process we came across various self-declared \textbf{textual types} – often within a single web domain –, we also established a \textit{source typology}, formed by the sources’ classification of their own texts. Therefore, the \textit{broad typology} contrasts with the source typology: the former is an initial grouping by similar web-domain content, and the latter, usually a more detailed label for the types of texts provided by the source. An example would be a ``request for proposals" document, extracted from one of the websites of the Brazilian judicial branch. It has ``judicial branch" as its broad typology and ``request for proposals" as its source typology, according to what was explicitly stated on the website from which the document was extracted.

In addition, we defined a third categorization of our sources, the \textit{domain} metadata category, which classifies the social environment in which these textual types occur, e.g., academic, entertainment, pedagogical. In the ``request for proposals" example, in the previous paragraph, the domain is ``judicial".

After downloading and processing the texts, they are embedded in an XML \textbf{header} with multiple categories of metadata. The Carolina header was defined in accordance with the “Corpus” customization provided by the TEI Consortium (Text Encoding Initiative) Guidelines (TEI Consortium 2021) \citep{tei2021}. TEI is a modular and flexible system, whose infrastructure enables users to create a specified encoding schema appropriate to their needs without compromising data interoperability.

The \textbf{metadata} is divided in two categories: \textit{File Description}, which groups technical information of the file generated for the corpus and its processing, such as extraction date and researcher responsible for the metadata; and \textit{Source Description}, which includes information from the source document, such as license and file extension.\footnote{An example of a text embedded in the header with its metadata can be accessed at \url{https://sites.usp.br/corpuscarolina/exemplo/}}

In the appendix we present two tables that show the metadata information and its description. As we process a large volume of texts, most metadata information cannot always be retrieved. The majority of the header categories are optional, and only the information that can at all times be automatically obtained is mandatory.

\section{Current State}
\label{sec:current}

The corpus current version, Ada 1.1, \footnote{All future 1.x versions will be named after Ada Lovelace (1815-1852), who wrote the first programming algorithm.} is available at Portulan Clarin, \footnote{\url{https://portulanclarin.net/repository/browse/carolina-general-corpus-of-contemporary-brazilian-portuguese-with-provenance-and-typology-information/f3751b34e36611ecaa5802420a870112f00a37650c304dbda703d85e14a2e945/}} a platform to make linguistic resources accessible to researchers. It can also be accessed at Hugging Face, \footnote{\url{https://huggingface.co/datasets/carolina-c4ai/corpus-carolina}} a platform for AI resources, which is a convenient tool for developing NLP applications, providing easy ways to manipulate corpus data, although it requires some programming knowledge.

Regarding the language variants of the corpus, the texts are mainly in Brazilian Portuguese, apart from some web domains that do not distinguish between variants, i.e. Twitter and Wikipedia. During the surveys, however, we learned that the majority of Portuguese-speaking users in both domains are Brazilian, so we decided to maintain these sources. Even so, we kept their texts separate from the rest of the corpus, which is in Brazilian Portuguese, in order to allow greater control to the corpus user.

\begin{table}[!ht]
    \center
    \begin{tabular}{lrrrr}
        \toprule[1pt]
        \textbf{Broad typology} & \textbf{Files}    & \textbf{Size} & \textbf{Texts}    & \textbf{Tokens} \\
        \toprule[1pt]
        Datasets and other corpora  & 68    &  3.2GB    & 1,098,717   & 214,856,955   \\
        Judicial branch             & 29    &  1.4GB    &   38,068   & 195,145,400   \\
        Legislative branch          & 1     & 19.4MB    &      14   &   3,097,274   \\
        Public domain works         & 1     &  3.5MB    &      26   &     592,288   \\
        Social media                & 1     & 14.0MB    &    3,449   &   1,305,058   \\
        University domains          & 1     &  9.3MB    &     945   &   1,119,686   \\
        Wikis                       & 55    &  2.6GB    &  603,968   & 237,205,916   \\
        \midrule[.7pt]
        \textbf{Total}              & \textbf{156}  & \textbf{7.2GB}    & \textbf{1,745,187}  & \textbf{653,322,577} \\
        \end{tabular}
    \caption{Carolina version 1.1 (Ada) in numbers}
    \label{tab:carolina_numbers}
\end{table}

\begin{figure}[H]
    \centering
    \includegraphics[scale=.2]{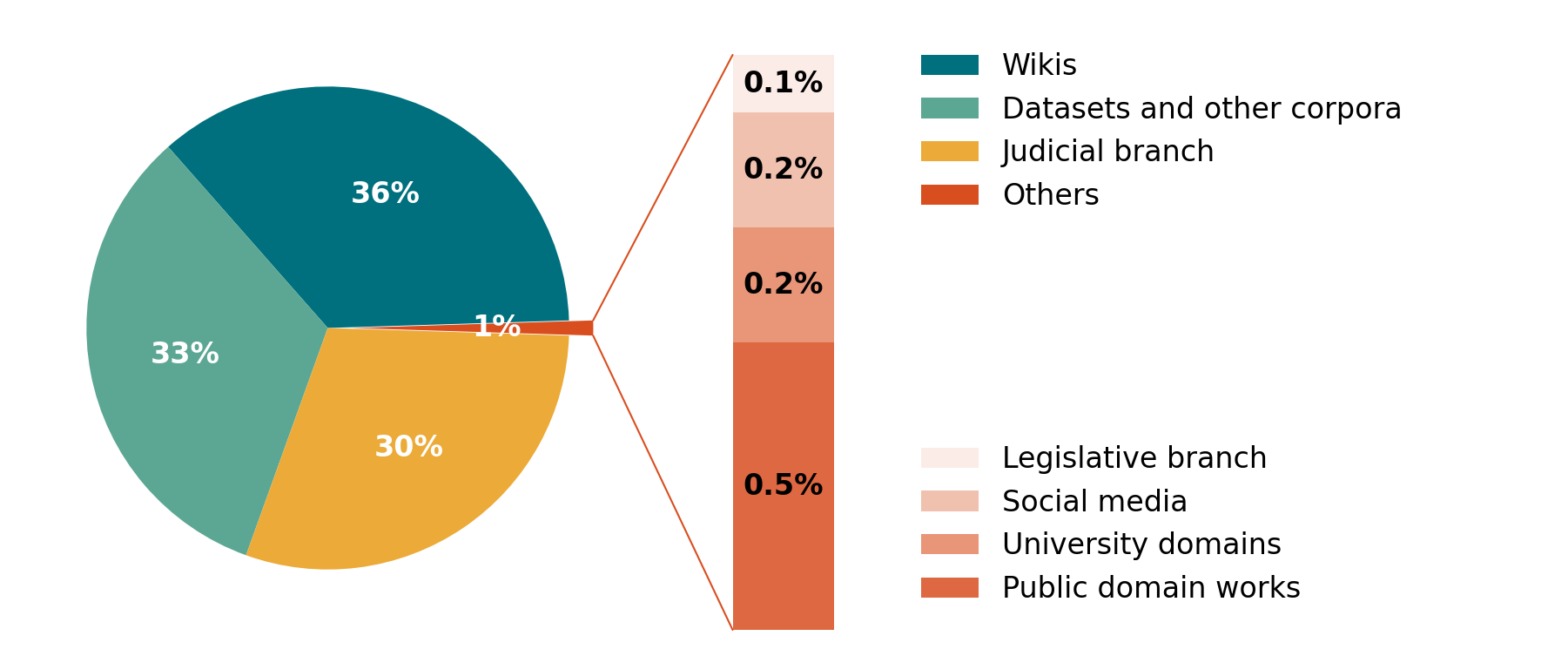}
    \caption{Percentage of tokens by broad type.}
    \label{fig:broad}
\end{figure}

\begin{figure}[H]
    \centering
    \includegraphics[scale=.2]{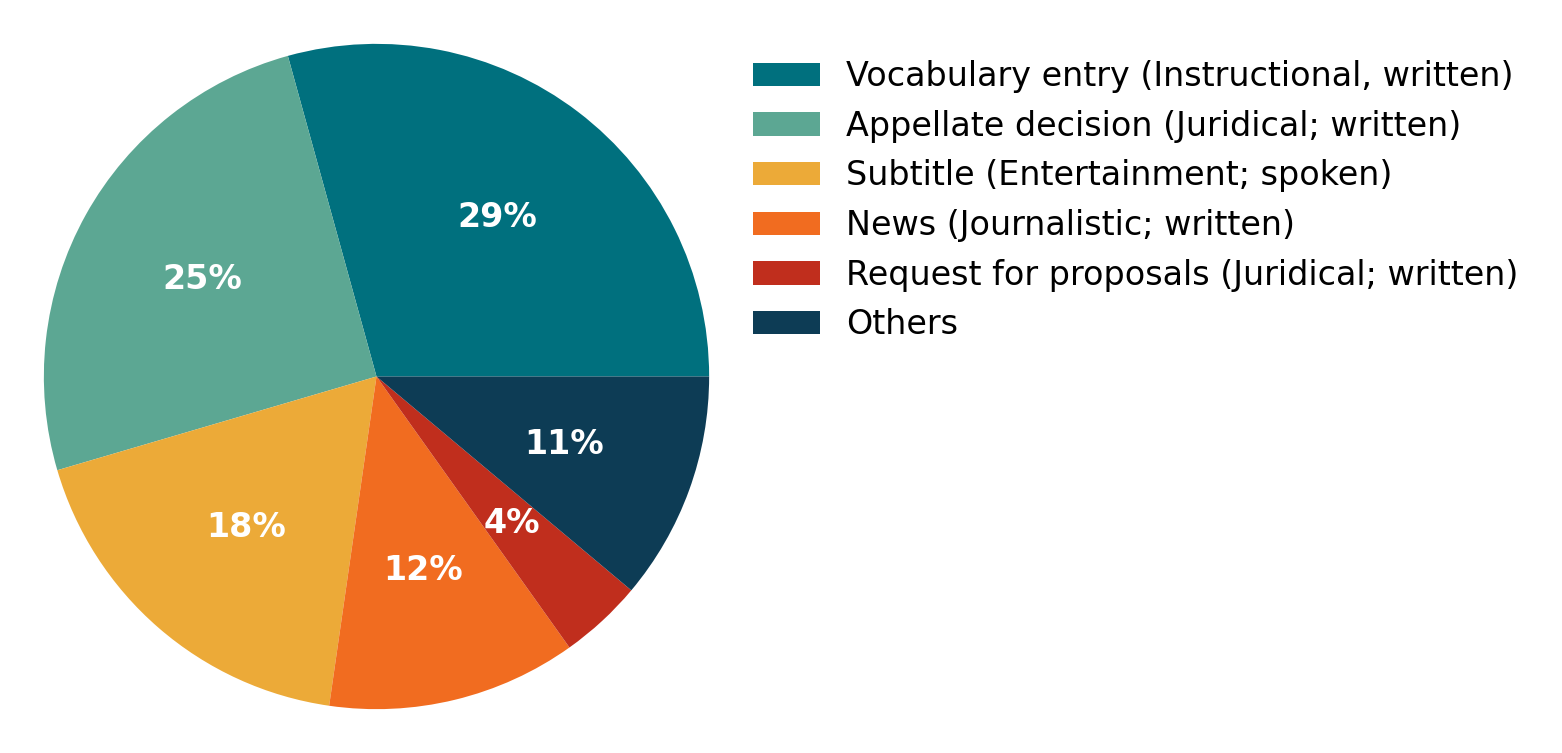}
    \caption{Percentage of tokens by source type.}
    \label{fig:source}
\end{figure}

\begin{figure}[H]
    \centering
    \includegraphics[scale=.2]{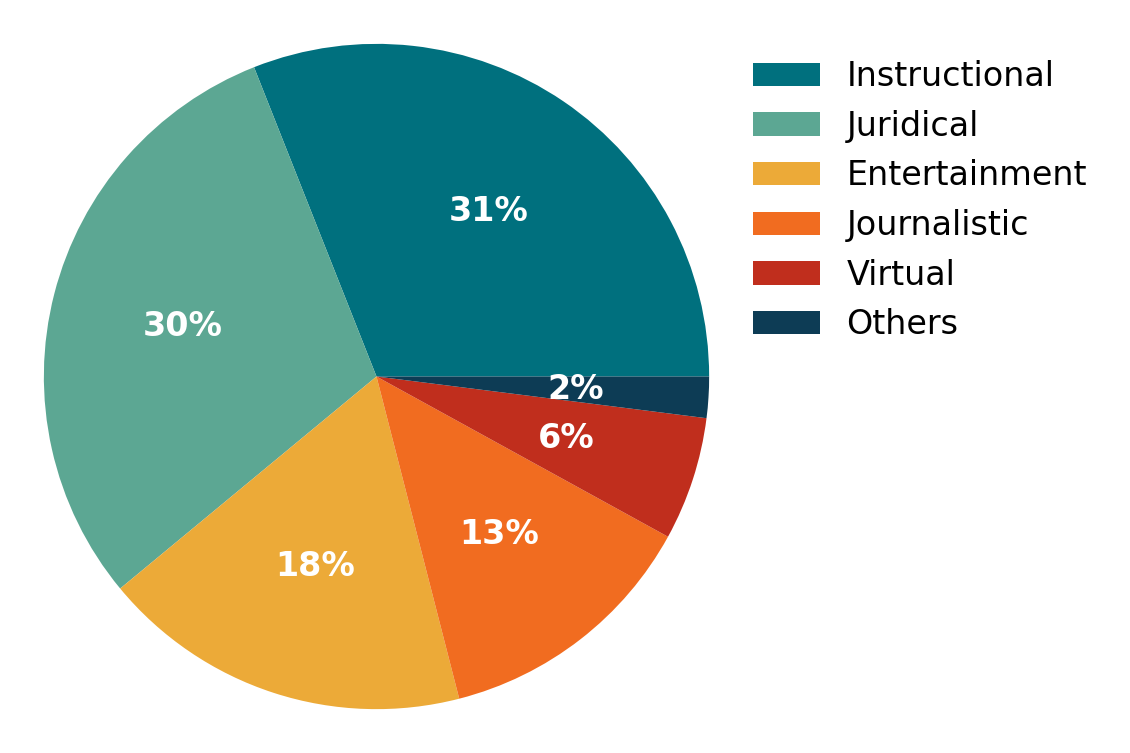}
    \caption{Percentage of tokens by domain.}
    \label{fig:domain}
\end{figure}

As far as text distribution is concerned, the corpus is well balanced considering its three major broad typologies: ``Wikis" (36 per cent), ``Datasets and other Corpora" (33 per cent) and ``Judicial branch" (30 per cent), though it is not balanced when considering all six broad typologies (figure \ref{fig:broad}) or when considering the source typologies (figure \ref{fig:source}) and the domain tag (figure \ref{fig:domain}). This is also shown in Table \ref{tab:carolina_numbers}. However, this was to be expected, especially concerning the broad and the source typology, for the former is a general grouping made to guide our work, and the latter is how our sources classify their texts. As Carolina is web-based and so far restricted to the sources that allow free redistribution, our corpus balancing is limited by the distribution of the contents under these conditions. As for the domains, new surveys will be made to ensure better balancing in future versions. We also intend to create balanced sub-corpora, assembling smaller existing parts of broad and source types.\footnote{Here, we are referring to an equal balance between typologies and not to a faithful depiction of what is observed on the internet. It would be difficult to gauge the parameters of such balancing and even more difficult to stick to it given the restrictions of open licensing.}

Regarding texts' licenses, the corpus has documents that were originally distributed under $32$ different licenses, the most common being: \texttt{CC BY-NC-SA 4.0}, \texttt{CC BY-SA 3.0}, and \texttt{CC0: Public Domain}. To facilitate the corpus use, we recast the majority of the texts into two compatible different licenses: \texttt{CC BY-NC-SA 4.0}\footnote{\url{https://creativecommons.org/licenses/by-nc-sa/4.0/}} and \texttt{CC BY-SA 4.0}.\footnote{\url{https://creativecommons.org/licenses/by-sa/4.0/}} In this process, each corpus entry was attributed to a license equivalent to or more restrictive than its original license, ensuring its correct distribution.

In the current version, the \texttt{CC BY-NC-SA} license was attributed to $66.78$ per cent of the texts and the \texttt{CC BY-SA} license to $33.22$ per cent. One document, distributed under GNU, was not compatible with the recasting, maintaining its original license.

In all cases where the recasting was possible, the new license is expressed in the tag ``License" under the ``File Description" category of its metadata section. However, information about the original license of each document is maintained, available in the tag ``License" as well, but under the ``Source Description" category of the metadata.

\section{Derivative Works}
\label{sec:derivative}

Carolina was developed by a multidisciplinary team of computer scientists and linguists, with the intention of producing a multi-purpose corpus for both research fields. We hope that those interested will be able to use Carolina for various applications, as the corpus is openly available with the licenses for each segment carefully described. 

Our team has already started using Carolina to explore various scientific questions. In this section, we list ongoing applications, with the intention of inspiring readers to develop their own uses and methodologies.

In the field of Computer Science, two investigations conducted by graduated students are already under development: (i) one is training large-scale language models based on Transformers for Brazilian Portuguese. These models are using the data of Carolina as part of its training set. The same project intends to experiment with the division that Carolina offers by typologies to explore training modules focused on specific typologies, that can be later combined to generate more powerful language models; (ii) the other investigation is focused on the task of natural language inference (NLI), which intends to mine inferences of different typologies from Carolina, to be used in the construction of a new dataset for the NLI in Brazilian Portuguese.

In the field of Linguistics, our team is developing a research that seeks to investigate the linguistic aspects of different typologies, as well as the possibilities that the various fields of annotated metadata in Carolina offer, and their implications for linguistic theory and application.

Our team is also focusing efforts on disseminating the corpus within the NLP community and exploring it as a teaching tool for Computational Linguistics and Natural Language Processing. In 2023 we will offer two courses where the corpus will be employed: (i) the ``Natural Language Processing Laboratory" summer course at IME-USP, that will present the corpus as a tool for its students, locating it within the corpora scenario; (ii) the ``Computational Linguistics" discipline of the IME-USP Computer Science Postgraduate Course, that will use the corpus in application activities. We hope that such initiatives will contribute to introducing and familiarizing the Brazilian NLP community with Carolina, as well as starting to explore its didactic potential.

These, however, are just some of the many possible applications for Carolina, which our team will not be able to fully explore, but which we hope the Portuguese NLP community will.

\section{Conclusion and Future Steps}
\label{sec:future}

The Carolina Corpus is continuously being developed with the WaC-wiPT methodology (Web as Corpus with Provenance and Typology information), which allows its four distinguishing features: provenance, typology, text integrality and versioning. Although these features are individually present in other corpora, Carolina combines all of them, increasing the scope of uses for the corpus.

The Carolina Corpus has already been and will continue to be made available to the public according to a versioning plan. The following versions will focus on  increasing the size and diversifying the range of typologies, as well as creating tools for textual query.


We invite our readers from both fields to also make use of the corpus for their own applications.

\section*{Acknowledgements}

This work was carried out at the Center for Artificial Intelligence (C4AI-USP), with support by the São Paulo Research Foundation (FAPESP grant \#2019/07665-4) and by the IBM Corporation. This work was financed in part by the Coordenação de Aperfeiçoamento de Pessoal de Nível Superior -- Brasil (CAPES) -- Finance Code 001. M. Finger received partial support from FAPESP 2020/06443-5 (SPIRA), 2014/12236-1 (Animals) and CNPq 303609/2018-4 (PQ), and C. Namiuti received partial support from Bahia Research Foundation (FAPESB 0007/2016, 0014/2016). The authors declare that there are no competing interests other than those listed above.

\bibliographystyle{apalike}
\bibliography{bibliography}

\pagebreak
\section*{Appendix}
\label{firstpage}

\begin{table}[!h]
\centering
\begin{adjustbox}{width=\textwidth}
\begin{tabular}{
    >{\raggedleft\arraybackslash}p{.35\textwidth}p{.65\textwidth}}
\toprule[1pt]
  \multicolumn{1}{c}{\textbf{Tag}}     &
  \multicolumn{1}{l}{\textbf{Description}}  \\
\toprule[1pt]
\xml{title type="main"}     &   Name of the corpus  \\
\xml{title type="sub"}      &   Version of the corpus to which a file belongs   \\
\xml{authority}             &   Team responsible for the corpus         \\
\xml{sourceDesc}            &   Global description of the sources used in the corpus and their licenses \\
\xml{projectDesc}           &   Brief description of the corpus project \\
\xml{taxonomy}              &   Structure of the typologies (Source and Carolina) used to classify texts. They are subdivided in categories (\xml{category}) and their descriptions (\xml{catDesc}) \\
\xml{title}                 &   Id of the text within the corpus. It is a composition of 3 capital letters (according to the Broad Typology) + a serial number    \\
\xml{respStmt}              &   Person responsible for either the download, metadata, or extraction of the text, as indicated by the element \xml{resp} \\
\xml{measure unit="tokens"} &   Number of tokens in the text, indicated in the attribute "quantity" \\
\xml{authority}             &   Team responsabile for the distribution of the corpus' text  \\
\xml{date}                  &   Date in which the text was downloaded (type="Download") and date of processing and incorporation of such text into the corpus (type="Extraction")   \\
\xml{availability}          &   Status of availability of the corpus' text. The corpus' texts are always freely distributed \\
\xml{license}               &   The name of the license of the corpus' text. The attribute “target” contains the license url \\
\bottomrule[1pt]
\end{tabular}
\end{adjustbox}
\caption{Identification of metadata for \textit{File Description} category}
\label{tab:corpus_metadata_carolina}
\end{table}

\begin{table}[!h]
\centering
\begin{adjustbox}{width=\textwidth}
\begin{tabular}{
    >{\raggedleft\arraybackslash}p{.35\textwidth}p{.65\textwidth}}
\toprule[1pt]
  \multicolumn{1}{c}{\textbf{Tag}}     &
  \multicolumn{1}{l}{\textbf{Description}}  \\
\toprule[1pt]

\xml{name}                      & Original text title   \\
\xml{media}                     & Contains the attributes "mimeType", "url" and "source", all referring to the original file: "mimeType" refers to the file type, e.g. "text/csv", "application/pdf"; "source" contains the file name; and "url" refers to the file link    \\
\xml{author}                    & Author of the original file   \\
\xml{editor role="translator"}  & Translator of the original file   \\
\xml{sponsor}                   & Institution responsible for the original text \\
\xml{extent}                    & Original file size in bytes, tokens and pages. When the term “pages” does not apply to the text type, the quantity is “-1”    \\
\xml{publisher}                 & The publisher of the original text    \\
\xml{authority}                 & Authority responsible for the original text   \\
\xml{date}                      & Publishing date of the original text  \\
\xml{availability}              & Status of availability of the original text. Can be "free", when freely distributed, or "restricted", if a registration is needed to access the text  \\
\xml{license}                   & The name of the original text license. The attribute “target” contains the license url    \\
\xml{region}                    & Region of the linguistic variety  \\
\xml{seriesStmt}                & Contains the title of the whole work in the tag <title>, and the title of the work’s part in the tag <biblScope>, e.g. title of a story and title of a chapter    \\
\xml{sourceDesc}                & Description of the text origin, e.g. "Born digital"   \\
\xml{channel}                   & Attribute mode="w" for written texts, mode="s" for transcribed speech, or mode="m" for mixed  \\
\xml{constitution}              & Attribute type="single" for a complete text, type="frags" for a combination of incomplete texts, type="composite" for a combination of complete texts     \\
\xml{domain}                    & Social environment in which the textual types occur, e.g. academic, entertainment, pedagogical    \\
\xml{catRef}                    & Sources’ classification of their own texts. The attribute "scheme" indicates the Source Typology and "target" explicits the type (textual type + 3 capital letters of the domain + the first capital letter of the <channel> "mode")  \\
\xml{language}                  & Language variant of the text. The attibute "ident" contains "pt-BR", or "pt" if the variant is not specified  \\
\xml{catRef}                    & Initial grouping by similar web-domain content    \\
\xml{text}                      & Processed textual content \\

\bottomrule[1pt]
\end{tabular}
\end{adjustbox}
\caption{Identification of metadata for \textit{Source Description} category}
\label{tab:corpus_metadata_source}
\end{table}



   
 

\label{lastpage}

\end{document}